\documentclass[sigconf]{acmart}
\setcopyright{none}
\renewcommand\footnotetextcopyrightpermission[1]{}

\AtBeginDocument{%
  \providecommand\BibTeX{{%
    \normalfont B\kern-0.5em{\scshape i\kern-0.25em b}\kern-0.8em\TeX}}}

\setcopyright{none}
\copyrightyear{2021}
\acmYear{2021}
\acmDOI{10.1145/1122445.1122456}

\usepackage{amsmath}
\usepackage{amsfonts}
\usepackage{paralist}
\usepackage[subrefformat=parens]{subcaption}
\usepackage{arydshln}
\usepackage{comment}
\usepackage{url}
\usepackage{xcolor}
\usepackage{pifont}
\newcommand{\cmark}{\ding{51}}%

\newcommand{\Hline}{\noalign{\hrule height 1.0pt}} % 太い横線の定義

\usepackage{algorithm}
\usepackage{algpseudocode}
\algnewcommand\algorithmicinput{\textbf{Input:}}
\algnewcommand\INPUT{\item[\algorithmicinput]}
\algnewcommand\algorithmicoutput{\textbf{Output:}}
\algnewcommand\OUTPUT{\item[\algorithmicoutput]}
\algnewcommand\algorithmicforeach{\textbf{foreach}}
\algdef{S}[FOR]{ForEach}[1]{\algorithmicforeach\ #1\ \algorithmicdo}
\algnewcommand\algorithmicparallelforeach{\textbf{parallel foreach}}
\algdef{S}[FOR]{PForEach}[1]{\algorithmicparallelforeach\ #1\ \algorithmicdo}

\setcopyright{rightsretained}
\copyrightyear{2021}
\acmYear{2021}
\acmDOI{10.475/123_4}
\acmISBN{123-4567-24-567/08/06}
\acmConference[CIKM '21]{30th ACM International Conference on Information and Knowledge Management}{November 01--05, 2021}{Gold 
Coast, Australia}
\acmArticle{4}
\acmPrice{15.00}

\begin{document}

\title{FedMe: Federated Learning via Model Exchange}

%\author{Anonymous Submission \#1517}

\author{Koji Matsuda}
\email{matsuda.koji@ist.osaka-u.ac.jp}
\affiliation{
  \institution{Osaka University}
  \country{}
}
\author{Yuya Sasaki}
\email{sasaki@ist.osaka-u.ac.jp}
\affiliation{
  \institution{Osaka University}
  \country{}
}
\author{Chuan Xiao}
\email{chuanx@ist.osaka-u.ac.jp}
\affiliation{
  \institution{Osaka University}
  \country{}
}
\author{Makoto Onizuka}
\email{onizuka@ist.osaka-u.ac.jp}
\affiliation{
  \institution{Osaka University}
  \country{}
}

% \affiliation{%

% \affiliation{%

% \affiliation{%

% \affiliation{%

% \affiliation{%

% \affiliation{%

% \affiliation{%

\renewcommand{\shortauthors}{}

\begin{abstract}
% 1文目．対象領域の説明
% 2文目．その領域における現状技術の問題
% 3文目．上記の問題を解決する本論文の目標
% 4文目．提案技術の特徴
% 5文目．実験結果の要約

Federated learning is a distributed machine learning method in which a 
single server and multiple clients collaboratively build machine learning
models without sharing datasets on clients. 
%In general, federated learning aims to train a single global model for all clients, but the global model may not be optimal for all clients due to data heterogeneity. 
% Recently, it has received considerable attention from research communities 
% due to the privacy and network bandwidth concerns in many data-centric 
% applications (telecommunication, edge computing, IoT, etc.). 
Numerous methods have been proposed to cope with the data heterogeneity issue in federated learning. 
%One of the prevailing approaches is to build personalized models that are optimized for each client. 
Existing solutions require a model architecture tuned by the central server, yet a major technical challenge is that it is difficult to tune the model architecture due to the absence of local data on the central server. 
In this paper, we propose {\em Federated learning via Model exchange} 
(FedMe), which personalizes models with automatic model architecture tuning 
during the learning process. 
%The novelty of FedMe is its learning process; each client exchanges their personalized models with each other to train their own and other clients' models and tunes their model architectures based on the performance of other clients' models.
%To further improve the inference accuracy, we use two techniques; deep mutual learning and model clustering.
%Both techniques effectively support learning the models on clients.
The novelty of FedMe lies in its learning process: clients exchange their 
models for model architecture tuning and model training. 
First, to optimize the model architectures for local data, clients tune 
their own personalized models by comparing to exchanged models and picking 
the one that yields the best performance. 
Second, clients train both personalized models and exchanged models by using 
deep mutual learning, in spite of different model architectures across 
the clients. 
%In addition, we employ model clustering to select similar personalized models as exchanged models for each client, in order to  prevent models from overfitting the noise caused by deep mutual learning.
%In FedMe, clients can effectively build their personalized models with different architectures without sharing local data.
%In FedMe, clients can effectively build their personalized models with different architectures without sharing local data.
%Each client then trains models with two optimization techniques, deep mutual learning and model clustering, to improve inference accuracy.
%trains its own model, and the models are exchanged and trained for highly accurate personalized models. The accuracy of the personalized model can be improved by using deep mutual learning based on model clustering. 
%In addition, clients replace their own models with other clients' models if the other models are better than their own models.
%the model architecture can be dynamically determined by model replacement during each local training. 
We perform experiments on three real datasets and show that FedMe 
outperforms state-of-the-art federated learning methods while tuning 
model architectures automatically. 
\end{abstract}

\begin{comment}

\begin{CCSXML}
<ccs2012>
 <concept>
  <concept_id>10010520.10010553.10010562</concept_id>
  <concept_desc>Computer systems organization~Embedded systems</concept_desc>
  <concept_significance>500</concept_significance>
 </concept>
 <concept>
  <concept_id>10010520.10010575.10010755</concept_id>
  <concept_desc>Computer systems organization~Redundancy</concept_desc>
  <concept_significance>300</concept_significance>
 </concept>
 <concept>
  <concept_id>10010520.10010553.10010554</concept_id>
  <concept_desc>Computer systems organization~Robotics</concept_desc>
  <concept_significance>100</concept_significance>
 </concept>
 <concept>
  <concept_id>10003033.10003083.10003095</concept_id>
  <concept_desc>Networks~Network reliability</concept_desc>
  <concept_significance>100</concept_significance>
 </concept>
</ccs2012>
\end{CCSXML}

\ccsdesc[500]{Computer systems organization~Embedded systems}
\ccsdesc[300]{Computer systems organization~Redundancy}
\ccsdesc{Computer systems organization~Robotics}
\ccsdesc[100]{Networks~Network reliability}

\end{comment}


\maketitle
\pagestyle{plain}

\section{Introduction}
\label{sec:1_introduction}

%Motivation for Federated Learning, Description
With the growing popularity of mobile devices such as smartphones and tablets, 
an unprecedented amount of personal data has been generated. 
%Such personal data are useful to build machine learning models on many applications such as next-word prediction based on keyboard input history~\cite{hard2018federated}, wake word detection based on voice data~\cite{leroy2019federated}, and action recognition based on accelerometers and gyroscopes~\cite{Anguita2013APD}.
Such personal data are helpful to build machine learning models on a variety 
of applications such as action recognition~\cite{Anguita2013APD}, next-word prediction~\cite{hard2018federated}, and wake 
word detection~\cite{leroy2019federated}. 
%However, due to privacy and communication issues, it is not practical that a server collects all local data on clients and train models centrally.
However, due to the concerns raised by data privacy and network bandwidth limitation, it is impractical to collect all local data from clients and train 
models in a centralized manner. 
%To address these issues, {\em federated learning} was proposed~\cite{mcmahan2017communication},
%which is a decentralized learning mechanism to build models without sharing local data on clients.
To address the privacy concerns and network bandwidth bottleneck, {\em federated learning} has emerged as a decentralized learning paradigm to build a model without sharing local data on clients~\cite{mcmahan2017communication}.

%Explanation of Federated Learning
Federated learning builds a model with a single server and multiple clients in a 
collaborative manner. Its general procedure consists of two steps: ($1$) client 
learning, in which clients train models on their local data and send their 
trained models to the server, and ($2$) model aggregation, in which the server 
aggregates those models to build a global model and distributes the global model 
to the clients. These two steps are repeated until the global model converges. This 
procedure effectively uses clients' local data by sharing their trained models.
%Furthermore, federated learning protects the privacy of clients because they do not send their own local data~\cite{mothukuri2021survey}.

% \begin{table*}[!t]
% \caption{Comparison between FedMe and existing federated learning methods. \cmark{} and \xmark{} indicate satisfy and not, respectively. Hetero and Homo indicate that clients can have models with different model architectures and not, respectively.}
% \vspace{-3mm}
% \label{comparison}
% {\small
% \begin{tabular}{lcccccccccc}\Hline
%  & FedAvg  & HypCluster & MAPPER & pFedMe &  FML  & FedMD & FedNAS & Ours \\\hline
% Data heterogeneity & \xmark  & \cmark & \cmark & \cmark  & \cmark & \cmark & \xmark & \cmark \\
% Personalization & \xmark  & Homo & Homo & Homo &  Hetero & Hetero & \xmark & Hetero \\
% %Model heterogeneity & \xmark & \xmark & \xmark & \xmark & \xmark & \xmark & \xmark & \cmark & \cmark & \cmark \\
% Model architecture tuning & \xmark  & \xmark & \xmark & \xmark &  \xmark & \xmark & \cmark & \cmark\\\Hline
% \end{tabular}
% }
% \end{table*}

%Challenges of Federated Learning
%\smallskip
\noindent
{\bf A challenge of federated learning.} 
One of the challenges in federated learning is on data heterogeneity:
clients have local data that follow different distributions, i.e., they 
do not conform to the property of independent and identically distributed 
(IID) random variables. This 
causes difficulty in learning a single global model that is optimal 
for each client. Indeed, it has been reported that, in typical federated 
learning methods, model parameters of a global model are divergent when 
each client has non-IID local data~\cite{MLSYS2020_38af8613,li2019convergence}. 
%However, while data heterogeneity is harmful when creating a single global model, it is beneficial when creating models for each client. When creating models for each client, the accuracy is improved because the model only needs to fit the client's local data. As shown in Table~\ref{}, the test accuracy is higher when the data is heterogeneous than when the data is homogeneous.
%non-guaranteed convergence and model parameter divergence when a single global model is trained on non-IID data~\cite{hsieh2020non,li2019convergence,MLSYS2020_38af8613}.
Personalized federated learning methods have been proposed to deal with data  heterogeneity~\cite{mansour2020three,shen2020federated,NEURIPS2020_f4f1f13c,zhang2021personalized}.
These methods aim to build {\it personalized models}, which are optimized models for clients.

We have the following research questions for building optimal personalized models:

\begin{itemize}
\item %Data heterogeneity causes another challenge: 
{\it How to determine the model architectures of personalized models?} 
%Another challenge in federated learning is the problem of determining the model architecture. 
%In general, it is known that the more significant the amount of data, the better it is to use a larger model.
In existing personalized federated learning methods, the server must tune model architectures in advance. 
%Since the server is unware of distributions of local data on clients, the server needs to train multiple models with different architectures to tune model architectures remotely.
Since the server is unaware of local data distributions on clients, the server needs to train multiple models with different architectures to tune model architectures remotely.
However, this process requires high communication costs between the server and clients, making it impractical.
Recently, an automatic architecture tuning method was proposed to automatically modify model 
architectures during learning process~\cite{FedNAS}. It tunes the 
architecture of the single global model by the server. 
%Therefore, to find an optimal model architecture, the server must select multiple architectures to tune hyperparameters manually. 
However, it is likely that the model architecture tuned by the server is not 
optimal for each client, and the server is unable to evaluate the accuracy of 
the tuned model by using the local data of the clients.
%Therefore, model architecture should be tuned by each client to build the optimal  architecture for each client,  
Therefore, each client should individually tune its model architecture, 
%Because of data heterogeneity, model architecture tuned by the server may not be optimal for each client, so model architecture should be tuned by each client. 
which may differ across clients due to the non-IID data  
%In fact, personalized model architectures are different if each client trains on only their local data 
(see Table~\ref{tab:model_select} in our experimental study).
%and tuned, so, it is require to train between models with different architectures. 

To the best of our knowledge, there are no personalized federated learning methods 
that can automatically tune the model architecture during the learning process.
Since each client is unaware of the local data on the other clients, we need means of leveraging other clients' models to tune model architectures. 
%In addition, two research challenges exist. 
%it is not immediately evident that clients can optimally tune their 
%model architectures by themselves. This raises two research challenges. 
% \item {\em How do clients automatically tune their optimal model architectures by themselves?}
% Since each client is unaware of the local data on the other clients, we need means of leveraging other clients' local data to tune model architectures. 

\item {\em How does each client leverage other clients' models with different architectures to improve its model accuracy?}
The server may not aggregate personalized models because their model architecture may differ across the clients.
It is not effective to rely on the aggregation of models for leveraging other clients' models.
So, we need additional means of leveraging local data and models with different architectures. 
%Since model architectures may differ across the clients, it is not effective to simply aggregate the models. So we need means of utilizing local data without aggregating models with different architectures. 
\end{itemize}

%Our research challenges are (1) {\em how we automatically tune optimal model architectures for clients} and (2) {\em how we can effectively train between models with different architectures}.

%Therefore,  %by creating the most optimal model for each client (personalized model) through federated learning 
%\cite{mansour2020three,shen2020federated,NEURIPS2020_f4f1f13c,zhang2021personalized}.
%there are methods to deal with data heterogeneity by creating the most optimal model for each client (personalized model) through federated learning \cite{mansour2020three,shen2020federated,NEURIPS2020_f4f1f13c,zhang2021personalized}.
%The data heterogeneity causes another challenge; how to determine the model architecture.
%Another challenge in federated learning is the problem of determining the model architecture. 
%In general, it is known that the more significant the amount of data, the better it is to use a larger model.
%In federated learning, it is difficult to determine the optimal model architecture because the amount of local data cannot be known in advance. 
%To the best of our knowledge, there are no personalized federated learning methods that support automatic selection of model architecture.

%To the best of our knowledge, only FedNAS~\cite{FedNAS} supports automatic selection of model architecture.
%There is a method \cite{FedNAS} that searches for the model architecture, but there is no method that simultaneously performs model personalization and auto-tuning of the model architecture.

%Explanation of the proposed method 1
%\smallskip
\noindent
{\bf Contributions.} 
In this paper, we propose a novel federated learning method, {\em federated learning via model exchange} (FedMe for short). 
We propose a notion of \emph{exchanged models}, i.e., each client can receive models sent from other clients. Then the 
clients are able to tune model architectures and train their models by utilizing the exchanged models. 
%The novelty of FedMe is its learning process: clients exchange their models for model architecture tuning and model training.
%We call models that are sent from other clients {\em exchanged model}.
The learning process of FedMe addresses the aforementioned research questions. 
First, clients tune their model architectures based on the performance of exchanged models. 
To optimize the model architecture for local data, each client compares its own personalized
model to the exchanged models and pick the one that yields the best performance. In this way, 
clients can automatically and autonomously modify their model architectures. %so that their models optimize their local data.
Second, clients train both their own and exchanged models to improve both models, and the 
server aggregates the trained models of the same clients. 
We use two techniques for model training: deep mutual learning~\cite{8578552} and model clustering.
Deep mutual learning is effective in simultaneously training two models by mimicking the 
outputs of the models regardless of model architecture. %We can naturally adopt deep mutual learning.
Model clustering selects similar personalized models as exchanged models for each client, which 
prevents models from overfitting the noise caused by deep mutual learning. In doing so, the 
aggregated models can reflect the local data on other clients because they are trained by using 
other clients' local data and the exchanged models.
We evaluate the performance of FedMe by comparing with state-of-the-art methods on three real datasets.
%, we evaluate the performance of FedMe compared with the state-of-the-art methods.
Our experiments show that FedMe achieves higher accuracy than state-of-the-art methods even if we 
manually tune these methods for their best model architecture.
%In addition, we find a new insight that fine-tuning is enough to build highly accurate personalized models on clients, which is not evaluated fairly in existing studies.
Another interesting takeaway of the evaluation is that traditional federated learning methods 
with fine-tuning can build highly accurate personalized models on clients, which is not evaluated fairly in existing studies. 
%We open our source codes (including existing methods) for further research\footnote{\texttt{https://github.com/OnizukaLab/FedMe}}.

%when the same model is used for all clients and is comparable to that of existing methods when automatic selection of model architecture.
%Our evaluation experiments show that the accuracy is higher than that of existing methods when the same model is used for all clients and is comparable to that of existing methods when automatic selection of model architecture.

%Explanation of the structure of this paper
%\smallskip
\noindent
{\bf Organization.} 
The remainder of this paper is organized as follows. In Section \ref{sec:2_relatedwork}, we review related work. 
In Section \ref{sec:3_problem_definition}, we define the problem. We then present our proposed method, FedMe, in 
Section \ref{sec:4_FedMe}, and report our empirical evaluation results in Section \ref{sec:5_experiment}. In 
Section \ref{sec:6_conclusion}, we summarize the paper and discuss future work.

\section{Related Work}
\label{sec:2_relatedwork}

The research on federated learning has been actively studied since McMahan et al. introduced federated learning~\cite{mcmahan2017communication}.
%Academic and industrial researchers study in various aspects of federated learning such as security~\cite{bhagoji2019analyzing}, industrial applications ~\cite{leroy2019federated},  and frameworks and libraries~\cite{chaoyanghe2020fedml}.
%communication costs \cite{reisizadeh2020fedpaq} and on security\cite{bhagoji2019analyzing,bagdasaryan2020backdoor}. 
%Other projects include the application of optimization methods such as Adam to federated learning\cite{reddi2020adaptive}, industrial applications of federated learning\cite{hard2018federated,leroy2019federated}, frameworks and libraries\cite{ beutel2020flower,chaoyanghe2020fedml}. 
Several survey papers summarize studies of federated learning~\cite{kairouz2019advances,lim2020federated,mothukuri2021survey}.

Numerous federated learning methods have been proposed recently.
Thus, we describe only typical methods due to the page limitation.
%~\cite{mcmahan2017communication,Wang2020Federated,liu2019edge,mansour2020three,NEURIPS2020_f4f1f13c,zhang2021personalized,shen2020federated,li2019fedmd,FedNAS}.
%,guha2019one,chen2021fedbe,MLSYS2020_38af8613,reddi2020adaptive,karimireddy2020scaffold,arivazhagan2019federated,smith2017federated,fallah2020personalized,khodak2019adaptive}.
We review federated learning methods from three points of view; (1) data heterogeneity, (2) personalization
%(3) model heterogeneity, which is that methods work in environments that clients have models with different architecture each other;
and (3) model architecture tuning.
Methods for data heterogeneity aim to appropriately build models in environments that clients have non-IID local data.
Methods for personalization aim to build optimal personalized models for each client. 
Personalization has two types, homogeneous and heterogeneous, in which model architectures of all personalized models are the same and different, respectively.  
Methods with model architecture aim to automatically tune the model architecture.
% means whether the method's goal is to prevent the loss of accuracy due to data heterogeneity.
% Personalization means whether the method's goal is to train a personalization model for each client.
% Model heterogeneity means whether the method can train models with different structures for each client. 
% Model architecture tuning means whether the method can automatically tune the model structure. 
%We summarize the characterization of representative methods in Table~\ref{comparison}.

The most basic method on federated learning is FedAvg~\cite{mcmahan2017communication}, which aggregates all trained models of clients by averaging their model parameters to build a single global model.
Because the accuracy of FedAvg decreases in data heterogeneity, 
%McMahan et al. proposed FedAvg\cite{mcmahan2017communication}, which performs model aggregation by simply averaging the trained models, and
many methods have extended FedAvg to deal with data heterogeneity such as FedMA~\cite{Wang2020Federated} and HierFAVG~\cite{liu2019edge}.
Although these methods try to build a single global model by aggregating trained models, %considering the difference between trained models.
%, for example, Wang et al. proposed FedMA~\cite{Wang2020Federated} which aggregates units on each layer according to the similarity of parameters.
%, and Liu et al. proposed HierFAVG~\cite{liu2019edge} which aggregates model parameters on edges before aggregating them in the server. 
it is difficult to achieve high accuracy only by the single model.

% On the other hand, when there is data heterogeneity, the performance is degraded\cite{li2019convergence,zhao2018federated}.
% To address this issue, FedProx\cite{MLSYS2020_38af8613} and SCAFFOLD\cite{karimireddy2020scaffold} have been proposed. In these methods, performance can be prevented from falling by solving client drift, where the trained model for each client deviates from the ideal. However, these methods aim to train a single model. When there is data heterogeneity, the highly accurate model for all clients will be different from the models that are highly accurate for each client. Therefore, training models for each client improve accuracy rather than training a single model.

Personalized federated learning methods have been proposed to build different models for each client~\cite{mansour2020three,NEURIPS2020_f4f1f13c,zhang2021personalized}. 
These methods can increase the accuracy compared with methods that only build the single global model.
We first reviews homogeneous personalization methods, which build personalized models with different parameters but their model architectures are the same. 
%ed federate learning methods have two types; homogeneous and heterogeneous.
%They are generally homogeneous personalization, that is, all personalized models are designed with the same model architectures. 
%Many studies that address data heterogeneity by train models with different parameters for each client are proposed; 
Mansour et al. proposed HypCluster and MAPPER~\cite{mansour2020three}. 
In HypCluster, the server prepares several global models and distributes them to clients. Clients train only the model that has the highest accuracy and sends it back to the server. Then, the server aggregates each trained model as new global models.  
%in some similar trains multiple models. 
%Each client selects the $1$ model with the least loss and trains the selected model. 
In MAPPER, clients compute the balancing weights of the global model and its trained model and then do a weighted sum of their parameters. 
%Arivazhagan et al. proposed FedPer~\cite{arivazhagan2019federated}, in which only some of the layers of the model (the base layer) are sent to the server to be trained by federated learning, and each client trains the remaining layers (the individual layers). 
%Smith et al. proposed MOCHA~\cite{smith2017federated} which uses multi-task learning. 
%Fallah et al. proposed Per-FedAvg~\cite{fallah2020personalized} and Khodak et al. proposed FedAvg with ARUBA~\cite{khodak2019adaptive}. These methods incorporate meta-learning into FedAvg. 
T. Dinh et al. proposed pFedMe~\cite{NEURIPS2020_f4f1f13c}, which builds global and local models by normalizing using the Moreau envelope function. 
%Zhang et al. proposed FedFomo~\cite{zhang2021personalized}, which builds personalized models by optimizing the weighted average of models. The difference between FedFomo and MAPPER is that MAPPER does weighted averages of personalized and global models, and FedFomo does weighted averages of personalized models on all clients. 
These homogeneous personalization methods require the same model architecture for all personalized models, they cannot personalize their model architectures.
We here note that these methods do not use fine-tuning (i.e., after finalizing the models on learning process, clients do not re-train the models by their local data), and also they do not compare simple methods with fine-tuning (e.g., FedAvg with fine-tuning). Our experiments show that most methods have lower accuracy than FedAvg with fine-tuning.

There are heterogeneous personalization methods that build personalized models with different parameters and architecture across clients~\cite{li2019fedmd,shen2020federated}.
Clients can choose arbitrary model architectures depending on the size of local data and their computation resources. 
Shen et al. proposed Federated Mutual Learning (FML)~\cite{shen2020federated}.
%, in which clients train their personalized models to mimic the global models by deep mutual learning.
The server in FML distributes the global model and clients train both of the global and their personalized models by deep mutual learning. We use the similar idea of FML on client training, but FedMe does not build the global model.
%FML uses deep mutual learning between the models of each client and the models created by the server and sends only the models created by the server to the server for aggregation. 
%Since, in FML,  each client trains its model on its local data, the accuracy decreases when the local data is small.
Li et al. proposed FedMD~\cite{li2019fedmd} which incorporates knowledge distillation into federated learning.
FedMD needs public data, which is datasets similar to local data and can be used by all server and clients.
%FedMD consists of two stages: ($1$) transfer learning, where each client's model is trained on public data and then trained on local dataset, and ($2$) knowledge distillation, where the outputs of the client's model on public data are sent to the server and each client train its model with the averaged output as the correct label. Public data refers to data that all clients and servers can access; in FedMD, when the number of public data is small, accuracy decreases. 
%Also, FedMD needs public data. 
%Collecting data from clients leads to privacy issues, so the server needs to collect public data itself at a cost.
These heterogeneous personalization methods can build personalized models with different architectures for each client.
However, the server and clients need to determine model architectures before the learning process.
In addition, in our experiments, these methods cannot achieve higher accuracy than non-personalized methods.

Model architecture search is a hot topic in deep learning fields, which searches the best model architecture among predefined search spaces (e.g., layer types and the maximum number of layers)~\cite{he2020milenas,NAS}.
FedNAS supports the network architecture search on federated learning~\cite{FedNAS}. Although it automatically tunes model architecture, it aims to build a single global model. It does not aim to build personalized models. 

In summary, our method FedMe is the first method that can satisfy all the data heterogeneity, heterogeneous personalization, and model architecture tuning.

\section{Problem Definition}
\label{sec:3_problem_definition}
%Explanation Problem Defnition and Notation
In this section, we describe our problem definition. The notation used in this paper 
is summarized in Table~\ref{notation}.

Given a classification task, a server and a set of clients collaboratively build 
personalized models of clients. Let $S$ denote the set of clients. The number 
of clients is denoted by $|S|$. We use a subscript $i$ for the index of the 
$i$-th client. For example, $D_i$ is the local data of client $i$, and $n_i$ 
is the size of $D_i$ (i.e., the number of records). $N$ denotes the sum of 
$n_i$ for all clients. $x$ and $y$ are features and labels of records 
contained in the local data, respectively.
We assume classification tasks, so $y$ is assigned with a single class among $M$ classes. 
$T$ and $E$ are the total numbers of global communication rounds 
and local training rounds, respectively. Global communication means communication 
between the server and clients during training. Local training means that clients 
train the model on its local data. $t$ denotes an index of global communication 
round. 
%$E$ and $e$ are the total number and index of local training rounds per global communication rounds. 
$w^{t}_{{p}_i}$ and $w^{t}_{{ex}_i}$ are the personalized and exchanged model of 
client $i$ in round $t$, respectively. 
$Idx(w^{t}_{{ex}_i})$ is an index of the original client of $w^{t}_{{ex}_i}$. 
For instance, $Idx(w^{t}_{{ex}_i})$ returns $j$ if $w^{t}_{{ex}_i}$ is a personalized model of client $j$.

\begin{table}[!t]
\caption{Summary of notation used in this paper.}
\vspace{-3mm}
{\small
\label{notation}
\scalebox{0.95}{
\begin{tabular}{ll}\Hline
Symbol & Description\\\hline
$S$&  a set of clients\\
$i$ & an index of clients \\
$D_i$& $i$th client’s local data\\
$n_i$ & the size of $D_i$ \\
$x_i$, $y_i$ & a feature and label sampled from $D_i$, resp. \\
$M$ & the number of classes\\ 
$T$ & the number of global communication rounds\\
$t$ & an index of global communication rounds\\
$E$ & the number of local training epochs\\
%$e$ & an index of local training epochs \\
$w^{t}_{{p}_i}$ & a personalized model of client $i$ at round $t$\\
$w^{t}_{{ex}_i}$ & an exchanged model of client $i$  at round $t$ \\
$Idx(w^{t}_{{ex}_i})$ & an index of the original client of $w^{t}_{{ex}_i}$ \\
$C_k$ & models in cluster k\\
$K^t$ & the number of clusters at round $t$\\
\Hline
\end{tabular}}
}
\end{table}

\begin{figure*} [!t]
    \centering
    \includegraphics[width=0.9\linewidth,pagebox=cropbox,clip]{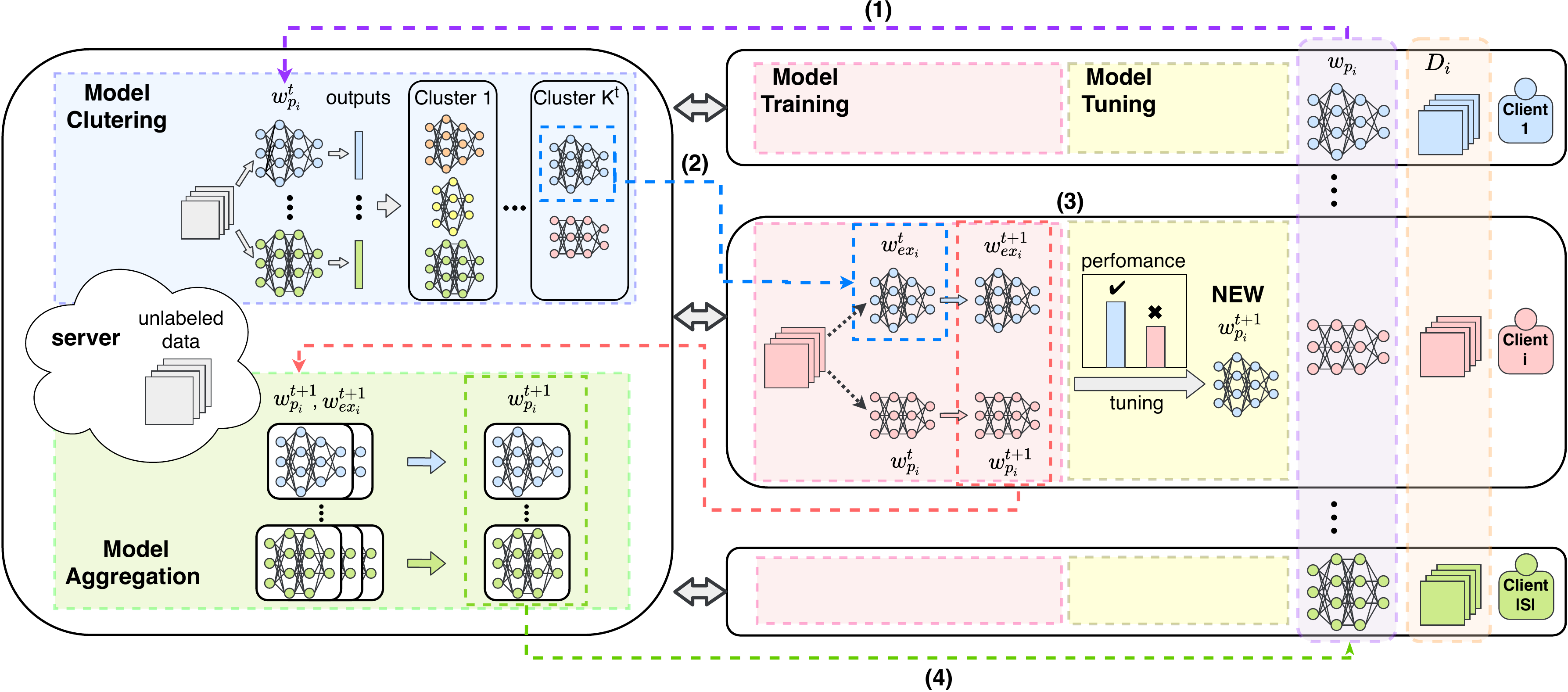}
    \caption{FedMe framework. 
    %In FedMe, ($1$) clusters models by their outputs on unlabeled data, ($2$) trains models by deep mutual learning, ($3$) determines which models to receive after aggregation, ($4$) average the model parameters for each personalized model, and return new personalized models from the server. These steps are repeated until the models converging.
    }
    \label{fig:FedMe}
\end{figure*}

In FedMe, each client builds its personalized model instead of a single global model. 
We define the optimization problem as follows:

\begin{gather}
  \{w_{p_1},\ldots,w_{p_{|S|}}\}=\text{argmin}\sum_{i\in |S|}\mathcal{T}_i(w_{p_i}). 
\end{gather}
\noindent
$\mathcal{T}_i$ is the personalized objective for client $i$, and is defined as follows: 
\begin{gather}
  \mathcal{T}_i = min\;\mathcal{L}(w_{p_i},D_i), 
\end{gather}
where $\mathcal{L}(w_{p_i},D_i)$ : $\Theta\longmapsto\mathbb{R}$ is the loss function of client $i$, corresponding to $D_i$ and $w_{p_i}$. $\Theta$ is the space of models, and is not fixed. 
This optimization problem is similar to that of~\cite{zhang2021personalized}.
In ~\cite{zhang2021personalized}, since the model architecture of personalized model is fixed, the size of $w_{p_i}$ is predetermined and fixed. In contrast, the size of $w_{p_i}$ in our problem is also optimized, which indicates that our problem aims to optimize the model architectures of personalized models. 
By solving this optimization problem, we can build optimal personalized models for each client.
\section{Methodology}
\label{sec:4_FedMe}

In this section, we describe our proposed method, FedMe. 
We first explain the overall idea and the framework of FedMe. 
After that, we present an algorithm and its technical components in detail. 
Finally, we shows a concrete example of our algorithm.

\subsection{Idea and Framework}
\label{overrall}
%FedMe aims to heterogeneous personalized federated learning with automatic model architecture turning. 
FedMe is a heterogeneous personalized federated learning method with automatic model architecture tuning.
Recall our research challenges; how we automatically tune optimal model architectures for clients and how clients use models with different architectures for improving their models.
For solving these research challenges, our idea is simple; clients receive models of other clients for leveraging model architecture tuning and send their models to other clients for training models by local data on other clients.
In other words, clients exchange their models for model architecture tuning and model training.

FedMe effectively leverages exchanged models by the following ways.
First, clients tune their personalized models based on the performance of exchanged models.
In more concretely, clients replace their personalized models with the exchanged models if the exchanged models have smaller loss on their local data than their personalized models.
Each client can automatically tune its model architectures so that the accuracy on its local data improves. 
Second, clients train both their personalized and exchanged models and the server aggregates the trained models of the same client.
This achieves model training that can train personalized models with different architectures.
Third, clients simultaneously and effectively train the both personalized and exchanged models by using deep mutual learning and model clustering.
Deep mutual learning simultaneously trains two models by mimicking outputs of models each other regardless of the model architecture.
The output of the other models may become noise and may overfit the noise when models are trained by significantly different local data~\cite{chen2021fedbe}. 
To prevent models from overfitting the noise, the server performs model clustering to select models with similar outputs. The model clustering groups model into subsets of models that have similar outputs by using Kmeans method~~\cite{macqueen1967some}.

Figure \ref{fig:FedMe} shows a framework of FedMe.
FedMe has five learning processes; (0) each client creates its personalized model with arbitrary model architectures, (1) clients send their personalized models to the server, (2) the server decides exchanged models for the clients based on model clustering and sends the exchanged models to clients, (3) clients train both their personalized and exchanged models by deep mutual learning and tune their personalized models based on the performance of the exchanged models, and (4) after clients send back the trained exchanged and personalized models, the server aggregates personalized and exchanged models for all clients and then send their aggregated personalized models to clients. FedMe repeats (1)--(4) until the number of global communication reaches a given threshold $T$.

\subsection{Algorithm}
\label{algorithms}
In this section, we describe the algorithm of learning procedures of FedMe.
The pseudo-code of FedMe is shown in Algorithm \ref{alg1}. After clients initialize their models (line $1$), FedMe starts its learning process. First, clients send their personalized model to the server (line $4$). The server clusters the models using unlabeled data (line $6$), and each client receives a model that belongs to the same cluster as an exchanged model from the server (lines $8$--$9$). Each client then trains the personalized and exchanged models (line $11$) and determines the index of its new personalized model (line $13$).
Each client sends two trained models and $a^t_i$ to the server (line $14$). The server aggregates each model by averaging their parameters (line $16$) and sends them to each client based on $a^t_i$ (line $18$). These steps are repeated until the number of global communication rounds becomes $T$. 

We explain detailed procedures of initialization, model training, model tuning, model clustering, and model aggregation in the following.

\begin{figure}[ht]
  \begin{algorithm}[H]
    \caption{Algorithm of FedMe.}
    \label{alg1}
    \begin{flushleft}
    {\bf Input}: number of global communication rounds $T$, number of local training epochs $E$, set of clients $S$ and their local data $\{D_i\}^{|S|}_1$, unlabeled data $U$, numbers of cluster \{$K^1,K^2,\cdots K^T$\}, learning rate $\eta$ \\
    {\bf Output}: personalized models $\{w^T_{p_i}\}^{|S|}_1$
    \end{flushleft}
    \begin{algorithmic}[1]
        \State \bf{Initialize($w^0_{p_i}$)} on all client $i$
        \For{$t=1,\cdots,T$}
            %\State $S_t \leftarrow$ (random set of $m$ clients)
            \For{$i\in S$}
                \State Client $i$ sends $w^{t-1}_{p_i}$ to server
            \EndFor
            \State $\{C_1,\ldots, C_{K^t}\}\leftarrow\bf{Model\;Clustering}$($\{w^{t-1}_{p_i}\}_{i\in S_t}$, $U$, $K^t$)
            %\State \hfill (subsection\ref{sec:model_clustering})
            \For{$i\in S$} 
                \State $w^{t-1}_{{ex}_i}\leftarrow w\in C_k\;that\;includes\;w^{t-1}_{p_i}$
                \State Server sends $w^{t-1}_{{ex}_i}$ to client $i$
                \For{$e=1,\cdots,E$}
                    \State $\bf{Model\;Training}$($w^{t-1}_{p_i}$, $w^{t-1}_{{ex}_i}$, $D_i$)
                    %\State \hfill (subsection\ref{sec:client_training})
                \EndFor
                \State $a^t_i\leftarrow\bf{Model\;Tuning}$($w^{t}_{p_i}$, $w^{t}_{{ex}_i}$, $D_i$)
                %\State \hfill (subsection\ref{sec:model_replacement})
                \State Client $i$ sends $w^{t}_{p_i},w^{t}_{{ex}_i},a^t_i$ to server
            \EndFor
                \State $\bf{Model\; aggregation}$($\{w^{t}_{p_i}$, $w^{t}_{{ex}_i}\}_{i\in S}$)
                %\State \hfill(subsection\ref{sec:model_aggregation})
            \For{$i\in {S}$}
                \State Server sends aggregated $w^{t}_{p_{a^t_i}}$ to client $i$
            \EndFor
        \EndFor
    \end{algorithmic}
  \end{algorithm}
\end{figure}

\noindent
{\bf Initialization.}
FedMe first requires the initialization of model architectures to clients.
Since clients can use arbitrary model architectures in FedMe, they can determine their model architecture depending on their local data.
For example, clients build optimal models on their local data.
Of course, the server can determine arbitrary models and distribute them to clients. 

\noindent
{\bf Model Training.}
%\label{sec:client_training}
In FedMe, each client exchanges its personalized model, and thus the model is trained on the local data of multiple clients, which enables training models even when clients have models with different architecture.
%while preventing the degradation of accuracy. 

Each client trains personalized and exchanged models on its local data by deep mutual learning.
Deep mutual learning between personalized and exchanged models improves accuracy compared to training them independently. 
Indeed, it is known that deep mutual learning effectively improves the inference performance of models when we use numerous models for training~\cite{8578552}.
Therefore, deep mutual learning has significant benefits on the learning process of FedMe.

%: its own personalized model and the exchanged model received from the server in the \ref{sec:model_clustering} .
% In order to increase the accuracy of each model over them trained independently, FedMe performs deep mutual learning for model training. 
% According to~\cite{8578552}, the more models we use, the more effectively we can train them by deep mutual learning.
% So, it is more effective in environments where a personalized model is created for each client. In addition, since FedMe performs model clustering in the \ref{sec:model_clustering} chapter, deep mutual learning can be performed on models with similar outputs.

We define loss functions $\mathcal{L}_{p}$ and $\mathcal{L}_{ex}$ of the personalized and exchanged models, respectively, as follows:
% In FedMe, two models, personalized and exchanged models, are trained by deep mutual learning. The loss functions of the two models are defined as follows: 
%\footnotesize
\begin{gather}
  \mathcal{L}_{p}\mathalpha{=}\mathalpha{-}\!\!\!\!\!\!\!\sum\limits_{(\!x,y\!)\in D_i}\!\sum\limits^M_{m=1}\!\!I(\!y,\!m\!)log(p^m_p(\!x\!)\!)\mathalpha{+}\!\!\!\!\!\!\sum\limits_{(\!x,y\!)\in D_i}\!\sum\limits^M_{m=1}\!p^m_{ex}(\!x\!)log\frac{p^m_{ex}(\!x\!)}{p^m_{p}(\!x\!)},\\
  \mathcal{L}_{ex}\mathalpha{=}\mathalpha{-}\!\!\!\!\!\!\!\sum\limits_{(\!x,y\!)\in D_i}\!\sum\limits^M_{m=1}\!\!I(\!y,\!m\!)log(p^m_{ex}(\!x\!)\!)\mathalpha{+}\!\!\!\!\!\!\sum\limits_{(\!x,y\!)\in D_i}\!\sum\limits^M_{m=1}\!p^m_{p}(\!x\!)log\frac{p^m_{p}(\!x\!)}{p^m_{ex}(\!x\!)},
\end{gather}
\normalsize
where $p^m_{p}$ and $p^m_{ex}$ are the predictions of the personalized and exchanged models for class $m$, respectively.
The first and second terms of these equations are the cross-entropy error and the Kullback Leibler (KL) divergence, respectively.
The function $I(y,x)$ returns 1 if $y=m$ and returns 0 otherwise. 

% , is defined as follows: 
% \begin{gather}
%     I(y_i,m)=\begin{cases}
%         1 & \text{if } y_i=m,\\
%         0 & \text{if } y_i\neq m.
%       \end{cases}
% \end{gather}

%$\mathcal{L}_{C_{p}}$ and $\mathcal{L}_{C_{ex}}$ are called the hard target loss of the personalized and exchanged models, respectively.
To minimize the above loss functions, client $i$ updates the two models.

\begin{gather}
  w^t_{p_i} \leftarrow w^{t-1}_{p_i} - \eta \nabla \mathcal{L}_{p},\\
  w^t_{{ex}_i} \leftarrow w^{t-1}_{{ex}_i} - \eta \nabla \mathcal{L}_{ex},
\end{gather}
\noindent
where $\eta$ is learning rate, and $\nabla \mathcal{L}_{p}$ and $\nabla \mathcal{L}_{ex}$ is gradient of personalized and exchanged models, respectively.

\noindent
{\bf Model Tuning.}
%\label{sec:model_replacement}
Clients can use models with any model architectures, which enabling clients to modify their models freely.
In the learning process of FedMe, each client has many opportunities to optimally modify its model because it receives models of other clients as exchanged models at each global communication round.
If the exchanged models achieve higher performance than the current models, they tune their models based on the exchanged models.

FedMe does not restrict means of model architecture tuning. 
In this paper, to validate the performance of design of FedMe, we use a simple tuning method which replaces their personalized models with the exchanged models. 
In more concretely, after each client trains models through deep mutual learning, it selects either its personalized or the exchanged models at round $t$.
FedMe computes $a^t_i$, which represents an index of personalized model that client $i$ selects, as follows:
%that  which client's model is received as the new personalized model after model aggregation.

\begin{gather}
    a^t_i = \begin{cases}
        i & \text{if } w^t_{p_i} \in \underset{w=w^t_{p_i},w^t_{{ex}_i}}{argmin}\mathcal{L}(w,D_i),\\
        Idx(w^t_{{ex}_i}) & \text{otherwise}.
      \end{cases}
\end{gather}

In this equation, each client compares the loss of the personalized and exchanged models and then replaces the personalized model with the exchanged model if the exchanged model has smaller loss than the personalized model. 
%$a^t_i$ represents which client's model is received as the new personalized model after model aggregation. 
%Since the model architecture with low accuracy is eliminated step by step, only the model architecture with high accuracy leaves to converge to the optimal.

Of course, we can use other tuning methods instead of replacements, for example increasing the number of layers.
Additionally, though we here consider the loss to tune the model architecture, each client can have its own criteria, such as the size of models and inference time.
We remain optimal model tuning methods on FedMe as future work.

% \begin{gather}
% a^t_i = \underset{w=w_{p_t},w_{{ex}_t}}{\min}\frac{1}{n_i}\sum\limits_{(x_i,y_i)\in D_i}f_i(x_i,y_i,w)
% \end{gather}

% %where $a^t_i$ is the index of the model with the lower loss in the global communication round $t$ among two models. Client $i$ receives a new personalized model based on $a^t_i$.
% %In this way, each client replaces the model architecture with one with high accuracy.
% Since the model architecture with low accuracy is eliminated step by step, only the model architecture with high accuracy leaves to converge to the optimal.

\noindent
{\bf Model Clustering.}
%\label{sec:model_clustering}
Due to the data heterogeneity among clients, the outputs of the personalized models differ among clients. 
If clients perform deep mutual learning between models with significantly different outputs, the models may overfit the noise.
%, resulting in inaccurate personalized models. 
In FedMe, models are clustered based on their outputs, and each client receives a model with similar output as an exchanged model from the server. 

Model clustering reduces the difference between the output of the own model and that of the other model, thus preventing overfitting the noise~\cite{gao2017knowledge}.
On the other hand, continuous training of models with similar outputs may lead less generality. 
Therefore, in the early stages of training, we do not perform model clustering to increase the generality of the models. As training progresses, we increase the number of clusters in the model clustering. 
In this way, the model can be personalized without overfitting while maintaining its generality.

Since federated learning does not share local data, we cannot use local data for model clustering. 
Therefore, FedMe assumes that the server has access to unlabeled data, such as one-shot federated learning ~\cite{guha2019one}, and uses unlabeled data $U$ as input.

We use the Kmeans method~\cite{macqueen1967some} to cluster models. 
%The personalized models of the participating client $i\in S_t$ are the targets of model clustering. First, the client $i\in S_t$ sends its personalized model to the server. 
The server first computes the outputs of the models using unlabeled data. 
The server then uses kmeans with those outputs and divides models into $K^t$ clusters.
% The output of the client $i$'s personalized model is $v_i$, the number of clusters in the global communication round $t$ is $K^t$, and the center point of cluster $j$ is $C_j$. 
% The loss function is defined as follows: 

% \begin{gather}
%   \mathcal{L}=\frac{1}{|S|}\sum^{K^t}_{j=1}\sum^{|S|}_{i=1}\delta^j_i ||v_i-c_j||^2,
% \end{gather}
% \noindent
% where, $\delta^j_i$ is the cluster assignment, which is defined by the following equation: 

% \begin{gather}
% \delta^j_i=\begin{cases}
%         1 & \text{if } j = argmin_k ||v_i-c_k||^2,\\
%         0 & \text{otherwise}.
%       \end{cases}
% \end{gather}

%Based on the cluster assignment, the server determines exchanged models. 
In the model exchange, each client receives a model of the same cluster as its own personalized model as exchanged models from the server at random. Here, if there is only one model in the cluster, the client receives a model from other clusters at random.

\noindent
{\bf Model Aggregation.}
%\label{sec:model_aggregation}
Client $i$ trains $w_{p_i}$ and $w_{{ex}_i}$ simultaneously. 
Therefore, it is necessary to aggregate all of them into a new model for each client.
FedMe aggregates the models by averaging the model parameters as in FedAvg.

\begin{gather}
  w^t_{p_i}\leftarrow \frac{1}{(s_i + 1)}(w^t_{p_i}+\sum^{|S|}_{j=1} u^t_{i,j} w^t_{{ex}_j}), 
\end{gather}
\noindent
where $s_i$ is the total number of clients that receive $w^t_{p_i}$ as the exchanged models. Also, $u_{i,j}$ represents which clients received the personalized model $w^t_{p_i}$ as an exchanged model and is defined by the following equation:

\begin{gather}
u^t_{i,j} = \begin{cases}
        1 & \text{if }i=Idx(w^{t}_{{ex}_j}),\\
        0 & \text{otherwise}.
      \end{cases}
\end{gather}

The model parameters are averaged and aggregated for each client's personalized model so that the aggregation is independent of the difference of model architecture.
%The new personalized model is returned to each client based on $a^t_i$ computed in the \ref{sec:model_replacement} chapter.

\subsection{Running Example}
\label{sec:example}

\begin{figure} [!t]
    \centering
    \includegraphics[width=0.9\linewidth,pagebox=cropbox,clip]{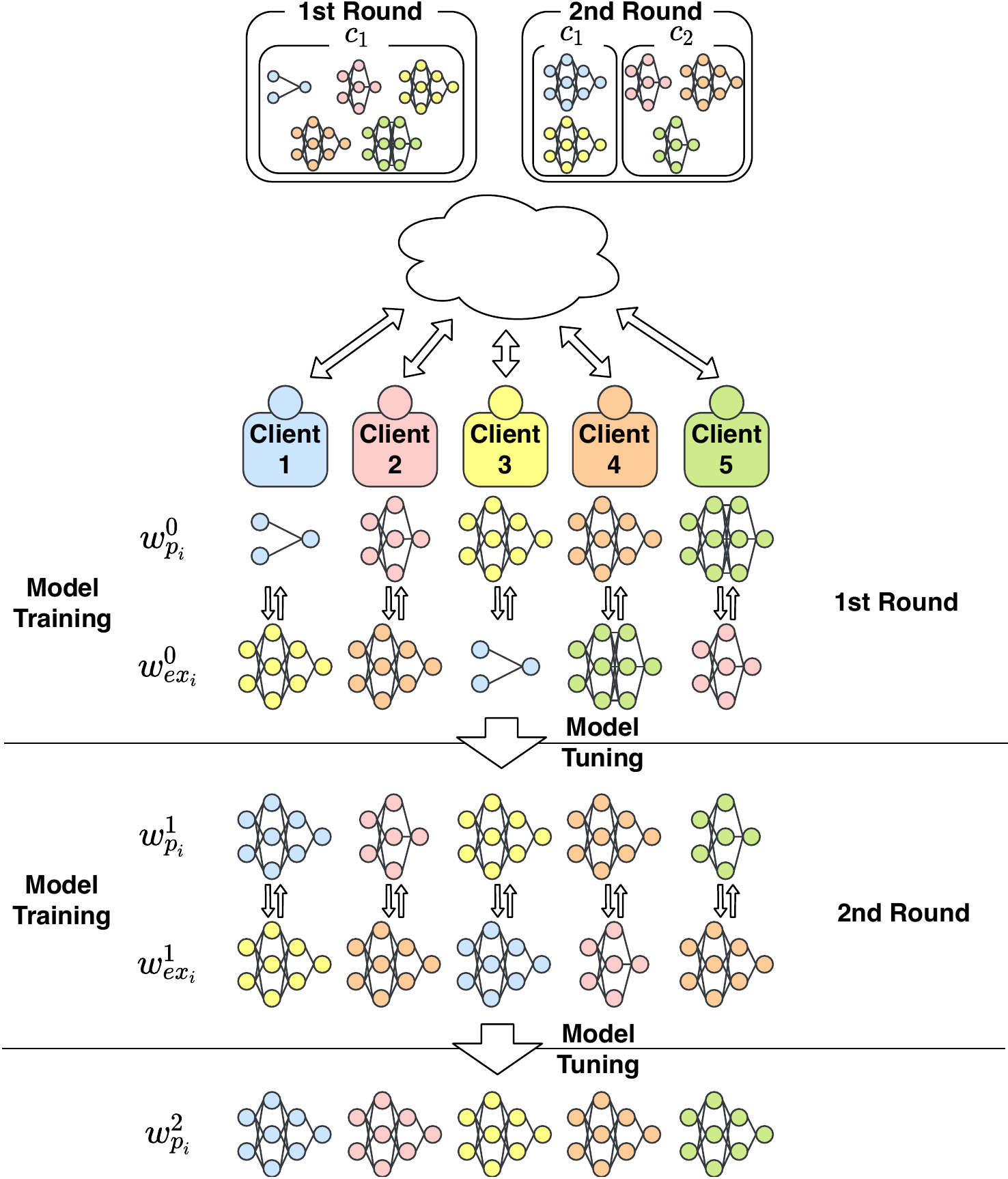}
    \caption{A running example of FedMe. 
    %In FedMe, ($1$) clusters models by their outputs on unlabeled data, ($2$) trains models by deep mutual learning, ($3$) determines which models to receive after aggregation, ($4$) average the model parameters for each personalized model, and return new personalized models from the server. These steps are repeated until the models converging.
    }
    \label{fig:FedMe_ex}
\end{figure}

We explain the FedMe algorithm using concrete examples. We assume that the number of clients is five and the number of global communication rounds is two. The number of clusters $K$ is initially one and increases by one at each global communication round.
Figure \ref{fig:FedMe_ex} illustrates the procedures of FedMe at the first and second global communication rounds.

\noindent
{\bf Initialization}: 
Clients initialized their personalized models.
In this example, each client selects model architectures depending on their local data and sets up each of these models as an initial personalized model as $\{w^{0}_{{p}_i}\}_{i=1-5}$.

\noindent
{\bf First round}: 
Clients send their personalized models to the server.
In the first global communication round, since the number of clusters is one, the server randomly selects exchanged models for each client.
In this example, clients $1$--$5$ receive $w^{0}_{p_3}$, $w^{0}_{p_4}$, $w^{0}_{p_1}$, $w^{0}_{p_5}$, and $w^{0}_{p_2}$ from the server as their exchanged models, respectively.

Each client trains its personalized and exchanged models by deep mutual learning, and update $\{w^{1}_{{p}_i},\;w^{1}_{{ex}_i}\}_{i=1-5}$, respectively.
Next, each client compares the loss of the two models on its local data. 
Suppose that the personalized model of clients $2$, $3$, and $4$ have smaller losses than their exchanged models, while the exchanged models of other clients have smaller losses. Thus, $\{a^1_i\}_{i=1-5}$ is $3$, $2$, $3$, $4$, and $2$, respectively.

All clients send the two trained models and $a^1_i$ to the server. 
The server then aggregates each personalized model. For example, the server aggregates $w^{1}_{p_1}$ and $w^{1}_{{ex}_3}$, $w^{1}_{{p_2}}$ by averaging their parameters. 
% aggregates $w^{1}_{p_2}$ and $w^{1}_{{ex}_5}$, and $w^{1}_{p_3}$ aggregates $w^{1}_{p_3}$ and $w^{1}_{p_1}$, $w^{1}_{p_4}$ aggregates $w^{1}_{p_4}$ and $w^{1}_{{ex}_2}$, and $w^{1}_{p_5}$ aggregates $w^{1}_{p_5}$ and $w^{t}_{{ex}_4}$ by averaging their parameters, respectively. 
The server sends back $w^{1}_{p_3}$, $w^{1}_{p_2}$, $w^{1}_{p_3}$, $w^{1}_{p_4}$, and $w^{1}_{p_2}$ to clients $1$--$5$, respectively, according to $a^1_i$. 
Each client sets up each of these models as a new personalized model, $\{w^{1}_{{p}_i}\}_{i=1-5}$.

\noindent
{\bf Second round}: 
We perform the second global communication round.
Clients send their personalized models to the server.
In the second global communication round, since the number of clusters is two, the server clusters the $5$ personalized models into two clusters using unlabeled data, and we assume that $\{ w^{1}_{p_1}, w^{1}_{p_3}\}$, and $\{ w^{1}_{p_2}, w^{1}_{p_4}, w^{1}_{p_5} \}$ belong to the same cluster, respectively. 
Clients $1$--$5$ receive $w^{1}_{p_3}$, $w^{1}_{p_4}$, $w^{1}_{p_1}$, $w^{1}_{p_2}$, and $w^{1}_{p_4}$ from the server as their exchanged models according to model clustering, respectively.

Then, each client trains its personalized and exchanged models by deep mutual learning and updates to $\{w^{2}_{{p}_i},\;w^{2}_{{ex}_i}\}_{i=1-5}$, respectively.
We assume that $\{a^2_i\}_{i=1-5}$ is $1$, $4$, $3$, $4$, and $4$ in the second round.

All clients send the two trained models and $a^2_i$ to the server. 
The server then aggregates each personalized model. 
% In more concretely, $w^{2}_{{p_1}}$ aggregates $w^{2}_{p_1}$ and $w^{2}_{{ex}_3}$, $w^{2}_{{p_2}}$ aggregates $w^{2}_{p_2}$ and $w^{2}_{{ex}_4}$, and $w^{2}_{p_3}$ aggregates $w^{2}_{p_3}$ and $w^{2}_{{ex}_1}$, $w^{2}_{p_4}$ aggregates $w^{2}_{p_4}$, $w^{2}_{{ex}_2}$ and $w^{2}_{{ex}_5}$, and $w^{2}_{p_5}$ aggregates only $w^{2}_{p_5}$ by averaging their parameters, respectively. 
The server sends back $w^{2}_{p_1}$, $w^{2}_{p_4}$, $w^{2}_{p_3}$, $w^{2}_{p_4}$, and $w^{2}_{p_4}$ to clients $1$--$5$, respectively, based on $a^2_i$, and each client sets up each of these models as its new personalized model.
$\{w^{2}_{{p}_i}\}_{i=1-5}$ are the final personalized models.

\section{Experiments}
\label{sec:5_experiment}
In this section, we test the accuracy of FedMe on three datasets with high degree of data heterogeneity. 
In our experiments, we aim to answer the following questions; 
\begin{description}
\item[Q1.] How accurate is the inference of FedMe compared with the state-of-the-art methods?
\item[Q2.] Does automatic model architecture tuning work well?
\item[Q3.] What techniques of FedMe impacts to the accuracy?
\item[Q4.] How fast is the learning process of FedMe compared with the state-of-the-art methods?
\item[Q5.] What is the impact of data heterogeneity and fine-tuning on FedMe?

%Experiments are conducted on two cases: ($1$) with pre-determined model architecture and ($2$) with automatic selection of model architecture. 
\end{description}
To simplify the experiments, we use Pytorch~\cite{paszke2019pytorch} to create a virtual client and server on a single GPU machine.

\subsection{Experimental Setup}
\label{sec:experiment_setting}
\subsubsection{Datasets, Tasks, and Models}
\label{model}

In the experiment, we use three settings; FEMNIST, CIFAR-10, and Shakespeare.
These datasets are frequently used in existing works~\cite{chen2021fedbe,li2019fedmd,MLSYS2020_38af8613,mansour2020three,mcmahan2017communication,Wang2020Federated}.
\begin{itemize}
\item FEMNIST: we use the Federated EMNIST-62 datset~\cite{caldas2018leaf}, which includes images of handwritten characters with $62$ labels. This dataset is divided into 3{,}400 sub data based on writers.
We conduct an image classification task.
\item CIFAR-10: We use CIFAR-10 dataset~\cite{cifar}, which includes photo images with 10 labels.
we divide the dataset into $20$ sub data using the Dirichlet distribution as in \cite{Wang2020Federated}.
We set two parameters $\alpha_{label}$ and $\alpha_{size}$ to decide the degree of heterogeneity of the size of local data and labels, respectively. We use 0.5 and 10 as $\alpha_{label}$ and $\alpha_{size}$, respectively.  
We conduct an image classification task. 
\item Shakespeare: We use Shakespeare dataset~\cite{MLSYS2020_38af8613}, which includes lines in ``The Complete Works of William Shakespeare''. This dataset is divided into 143 sub data based on actors.
We conduct a next-character prediction that infers next characters after given sentences. 
\end{itemize}
Table~\ref{tab:statistics} shows the statistics of the number of records on clients in datasets.
We here note that we randomly divide CIFAR-10 in each test, so the statistics of CIFAR-10 is an example value. 

We use different models for each setting following the existing works~\cite{reddi2020adaptive,Wang2020Federated}.
For FEMNIST and Shakespeare, we use CNN and LSTM, respectively~\cite{reddi2020adaptive}.
For CIFAR-10, we use VGG with the same modification reported in \cite{Wang2020Federated}.
In each setting, we use four models varying the number of layers.
For CNN and LSTM, we vary the number of convolution and LSTM layers from one to four, and the default value is two.
For VGG, we use VGG$11$, VGG$13$, VGG$16$, and VGG$19$, and the default is VGG$13$.

% For FEMNIST and Shakespeare, we use CNNs and LSTM with $x$ layers following~\cite{reddi2020adaptive}, respectively. We set $x$ as two as default.
% For CIFAR-10, we use VGG with the same modification of FedMA\footnote{\texttt{https://github.com/IBM/FedMA}}.

% The first two are image datasets, and we conduct the image classification. The last one is a text dataset., and we conduct the next-character prediction. CIFAR-10 consists of $50,000$ training and $10,000$ test data, and the statistics for FEMNIST and Shakespeare are shown in Table~\ref{tab:statistics}. For FEMNIST and Shakespeare, we use the same two convolutional layers of CNNs and two LSTM layers of RNNs as in~\cite{reddi2020adaptive}. For CIFAR-10, we use VGG-13 with the same modifications as FedMA\footnote{\texttt{https://github.com/IBM/FedMA}}.

\begin{table}[!t]
 \caption{Datasets Statistics.}
 \vspace{-3mm}
 \label{tab:statistics}
 \centering
 {\small 
 \begin{tabular}{ c c c c c c} \Hline
Datasets & Total num & Mean & STD & Max & Min\\\hline 
FEMNIST  &$671585$&$197.53$&$76.69$&$418$&$16$\\ 
CIFAR-10  &$49000$&$2450$&$1024.66$&$5018$&$1131$\\
%FEMNIST (test)  &$77483$&$22.79$&$8.51$&$47$&$3$\\  \hline
Shakespeare  &$413629$&$2892.51$&$5445.89$&$33044$&$2$\\ \Hline
%Shakespeare (test)  &$103477$&$723.62$&$1366.48$&$8261$&$1$\\ 
\end{tabular}
}
\end{table}

\subsubsection{Training and test}
In our experiments, the number of clients is 20.
In FEMNIST and Shakespeare, we select $20$ sub data for assigning local data on clients randomly.
In CIFAR-10, we randomly divide the whole data into 20 local data.
%and used them in the experiment. 
All clients participate in each global communication round following recent works~\cite{Wang2020Federated}.
We select 1{,}000 unlabeled data from each dataset. The unlabeled data was excluded from the train and test data.
We divide each local data into training and test data by the ratio of 9:1, 8:2, and 5:1 for FEMNIST, CIFAR-10, and Shakespeare, respectively.
Furthermore, we divide the training data into $7:3$ for FEMNIST and Shakespeare, and into $8:2$ for CIFAR-10, which are used as training and validation data, respectively.

% In FEMNIST and Shakespeare, we select $20$ clients randomly and used them in the experiment. In CIFAR-10, we divide the dataset into $20$ one using the Dirichlet distribution as in \cite{Wang2020Federated}, and use them as the local data. The Dirichlet distribution also determined the data size of each client, and the hyperparameters $\alpha_{label}$ and $\alpha_{size}$ is set to $0.5$ and $10$, respectively. For simplicity, we assume that all clients participate in each global communication round.
% The unlabeled data was $1000$ for all datasets, randomly selected from the remaining client data for FEMNIST and Shakespeare, and randomly selected from the $50000$ training data for CIFAR-10.
% For FEMNIST and Shakespeare, we divide the training data into $7:3$, and for CIFAR-10, divide into $8:2$, which were used as new training and validation data, respectively.

We set the number of global communication rounds to $300$, $500$, and $100$ for FEMNIST, CIFAR-10, and Shakespeare respectively, and set the local epoch $E$ to $2$ for all setting.
We conduct training and test five times and report mean and standard deviation (std) of accuracy over five times of experiments with different clients.

\subsubsection{Baselines and hyperparameter tuning}
We compare FedMe with three types of methods: ($1$) non-personalized federated learning methods, ($2$) personalized federated leaarning methods, and (3) non-federated learning methods.
For ($1$), we use FedAvg, and for ($2$), we use HypCluster, MAPPER, FML, and pFedMe.
For (3), we use Local Data Only, in which clients build their models on their model, and Centralized, in which a server collect local datasets from all clients (centralized can be considered as oracle).
We use fine-tuning on each client for Centralized, FedAvg, HypCluster, and FedMe after building their models.
In MAPPER and pFedMe, we do not use fine-tuning since their algorithms include the similar techniques to fine-tuning.
We implement all methods except for pFedMe\footnote{\texttt{https://github.com/CharlieDinh/pFedMe}} because these codes are not available.

%are fine-tuned with each client's local data.
%, and FML and pFedMe use personalized models for inference instead of global models.

%For reference, we also compare Local data only, where each client trains a personalized model using only its own local data, and Centralized, where the server trains a global model using all local data.

We explain hyperparameter tuning.
The learning rate is optimized for each method by grid search using a grid of $\eta\in\{10^{-3},10^{-2.5},$ 
$10^{-2},\ldots,10^{0.5}\}$. 
The optimization method is SGD (stochastic gradient descent) with momentum $0.9$ and weight decay $10^{-4}$. 
The batch sizes of FEMNIST, CIFAR-10, and Shakespeare are $20$, $40$, and $10$, respectively.
In Hypcluster, we use two global models.
In FedMe, we initialize model architectures of clients as the best accurate model on Local Data Only among model 1--4 (see Table~\ref{tab:model_select}).
We set the range of number of clusters to $1$--$4$; we initially use the number of clusters as $1$ and increase it by 1 at global communication round at $[150,225,275]$, $[250,375,450]$, and $[50,75,90]$ for FEMNIST, CIFAR-10, and Shakespeare, respectively.

\begin{table}[!t]
\caption{Average number of clients that select each model architecture based on their own local data. %Model 1--4 are 1--4 CNN layers in FEMNIST, (VGG$11$, VGG$13$, VGG$16$, and VGG$19$) in CIFAR-10, and 1--4 LSTM layers in Shakespeare, respectively.
}
\vspace{-3mm}
\label{tab:model_select}
\begin{tabular}{cccc}\Hline
 & FEMNIST & CIFAR-10 & Shakespeare \\\hline
model1 & $6.8\pm2.5$ & $5.6\pm1.5$ & $7.4\pm1.7$ \\
model2 & $7.4\pm0.5$ & $9.2\pm1.6$ & $6.6\pm1.8$ \\
model3 & $4.4\pm2.8$ & $4.0\pm2.0$ & $5.0\pm1.6$ \\
model4 & $1.4\pm1.1$ & $1.2\pm0.4$ & $1.0\pm1.7$\\\Hline
\end{tabular}
\end{table}

\subsection{Experimental Results}
\label{result}

\begin{table}[!t]
\centering
\caption{Test accuracy (mean$\pm$std).}
\label{tab:result_baseline}
\vspace{-3mm}
{\small 
\begin{tabular}{lccc}\Hline
 & FEMNIST & CIFAR-10 & Shakespeare \\\hline
Local Data Only & $64.71\pm2.94$ & $73.17\pm1.55$ & $24.77\pm1.95$ \\
Centralized & $79.35\pm 2.29$ & $90.80\pm0.92$ & $48.43\pm3.32$\\\hdashline
FedAvg & $77.25\pm 3.99$ & $89.59\pm0.94$ & $42.53\pm2.19$ \\
HypCluster & $76.29\pm3.15$ & $88.54\pm1.42$ & $41.10\pm3.29$ \\
MAPPER & $60.95\pm3.04$ & $61.29\pm4.19$ & $36.77\pm1.58$ \\
FML & $67.91\pm2.53$ & $79.89\pm1.44$ & $28.73\pm1.78$ \\
pFedMe & $72.92\pm3.54$ & $79.46\pm2.08$ & $40.33\pm2.27$ \\
FedMe & $\bf{78.52}\pm2.64$ & $\bf{89.76}\pm0.90$ & $\bf{44.71}\pm1.12$ \\\Hline
\end{tabular}
}
\end{table}

 \begin{figure*}[!t]
 \centering
    \begin{minipage}[t]{1.0\linewidth}
        \centering
        \includegraphics[width=0.50\linewidth]{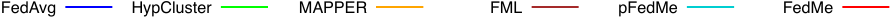}
    \end{minipage}
    \\
    \begin{minipage}[t]{0.30\linewidth}
        \centering
        \includegraphics[width=1.0\linewidth]{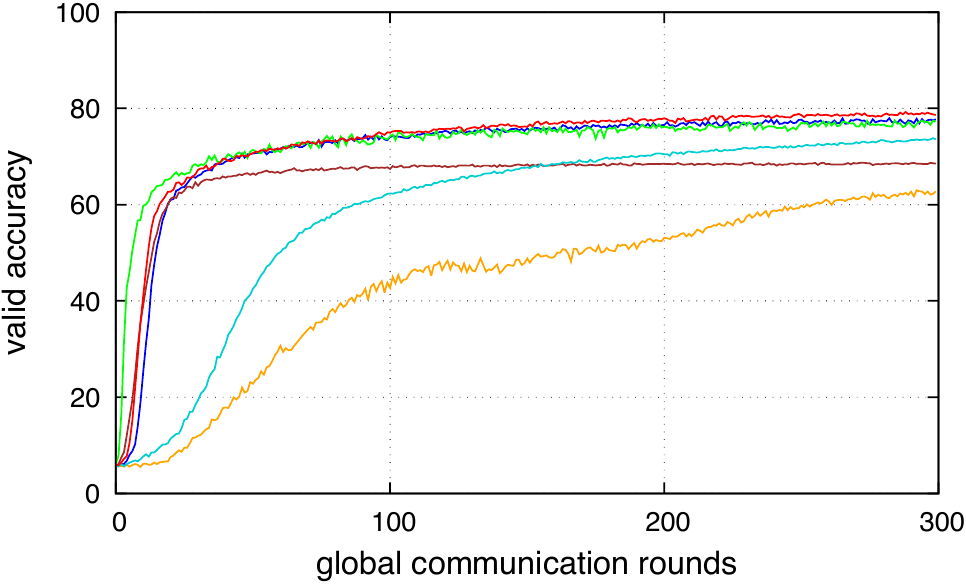}
        \subcaption{FEMNIST}
        \label{fig:femnist_result}
    \end{minipage}
    \begin{minipage}[t]{0.30\linewidth}
        \centering
        \includegraphics[width=1.0\linewidth]{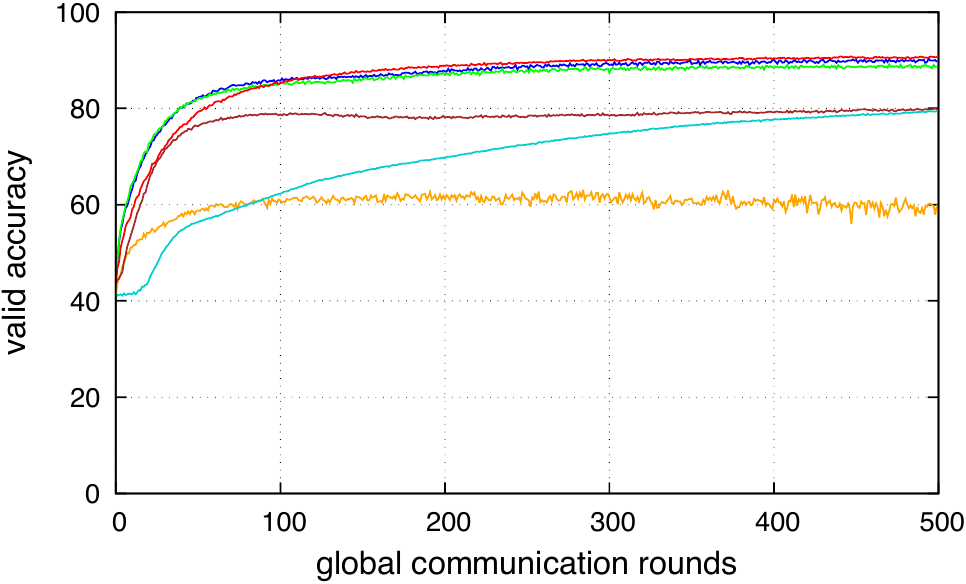}
        \subcaption{CIFAR-10}
        \label{fig:cifar10_result}
    \end{minipage}
    \begin{minipage}[t]{0.30\linewidth}
        \centering
        \includegraphics[width=1.0\linewidth]{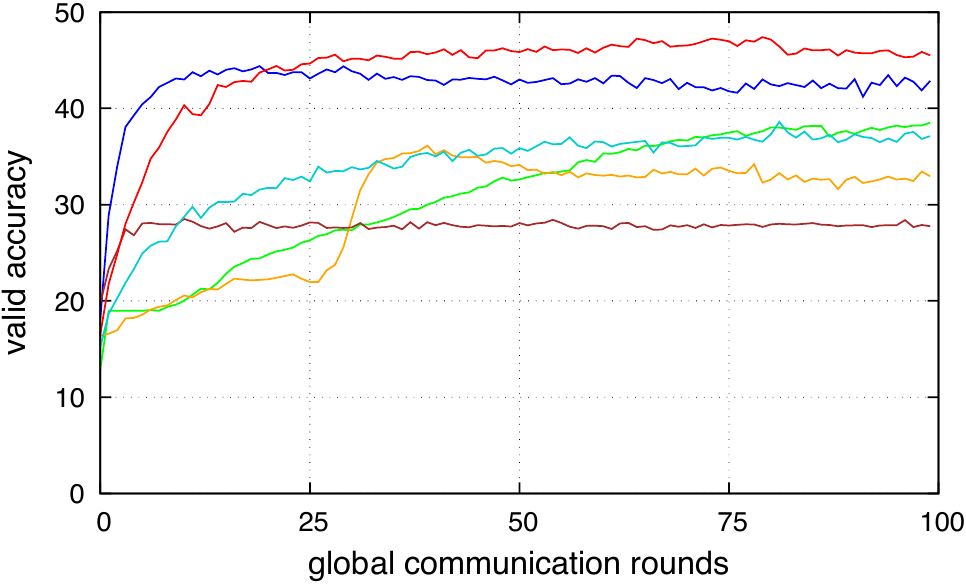}
        \subcaption{Shakespeare}
        \label{fig:shakespeare_result}
    \end{minipage}
\caption{The validation accuracy over time of various methods.}
\label{fig:result_valid}
\end{figure*}

%  \begin{figure}[!t]
%  \centering
%     \begin{minipage}[t]{1.0\linewidth}
%         \centering
%         \includegraphics[width=0.50\linewidth]{images/legend_data_time.pdf}
%     \end{minipage}
%     \\
%     \begin{minipage}[t]{0.45\linewidth}
%         \centering
%         \includegraphics[width=1.0\linewidth]{images/server_time.pdf}
%         \subcaption{FEMNIST}
%         \label{fig:server_time}
%     \end{minipage}
%     \begin{minipage}[t]{0.45\linewidth}
%         \centering
%         \includegraphics[width=1.0\linewidth]{images/client_time.pdf}
%         \subcaption{CIFAR-10}
%         \label{fig:client_time}
%     \end{minipage}
% \caption{Run time per global communication round}
% \label{fig:result_valid}
% \end{figure}

We show experimental results to answer the five questions.

\subsubsection{Q1. How accurate is the inference of FedMe compared with the state-of-the-art methods?}
Table~\ref{tab:result_baseline} and Figure~\ref{fig:result_valid} show the accuracy of FedMe and baselines.
Table~\ref{tab:result_baseline} shows average accuracy and standard deviation, and Figure~\ref{fig:result_valid} shows the validation accuracy at each global communication round. 

From Table~\ref{tab:result_baseline}, we can see that FedMe achieves the highest accuracy among federated learning methods for all setting and its accuracy is very close to accuracy of Centralized.
We here note that standard deviations of FEMNIST and Shakespeare are relatively large because clients differ in each test. FedMe achieves the lowerest (or the runner-up) standard deviation among federated learning methods for all settings, so we confirm that FedMe is the most robust among them.
This result shows that its learning process is effective.

Comparing the baselines, it is interesting in that FedAvg, which is the most simple method with fine-tuning, achieves the highest accuracy among baselines. 
This result indicates that data heterogeneity can be solved by fine tuning. 
We show more experiments related to data heterogeneity and fine tuning in Section~\ref{sec:experiment_Q4}.

From Figure~\ref{fig:result_valid}, we can see that FedAvg and FedMe generally have high accuracy at early rounds. 
This indicates that FedMe is early converge as the same as FedAvg.

%than the method that creates a personalized model for each client, indicating that the other existing methods are not effective for the three non-IID datasets in this experiment.
%Comparing FedMe with the existing methods, FedMe is the highest accuracy for all datasets, indicating that FedMe can create a more accurate personalized model than existing methods.

%Comparing the federated learning methods with Local Data Only, most federated learning methods are more accurate than Local Data Only. This indicates that Local Data Only causes over-fitting and that federated learning is effective. 

\subsubsection{Q2. Does automatic model architecture tuning work well?}
We here show how well FedMe tunes optimal model architecture automatically.
Table~\ref{tab_auto_selection} shows the accuracy of FedMe with fixed model architectures.
Model 1--4 are 1--4 CNN layers in FEMNIST, (VGG$11$, VGG$13$, VGG$16$, and VGG$19$) in CIFAR-10, and 1--4 LSTM laysers in Shakespeare, respectively.
%means that we initialize personalized models with the same architecture consisted of $1$--$4$ convolutional and LSTM layers for FEMNIST and Shakespeare, respectively, and with four models, VGG$11$, VGG$13$, VGG$16$, and VGG$19$ for CIFAR-10. 
For all setting, the accuracy of auto-tuning is middle among model 1--4.
In particular, in FEMNIST, the accuracy of auto-tuning is comparable to the highest accuracy of the pre-determined model architecture.
This result indicates that automatic model architecture tuning is effective without pre-defining the model architectures, so we can remove the cost to manually tune the model architecture.
%works well to decide at least, and depending on the dataset, works as well as manually tuning the model architecture.

\begin{table}[!t]
\caption{Impact of automatic model architecture tuning.}
\vspace{-3mm}
\label{tab_auto_selection}
\begin{tabular}{lccc}\Hline
 & FEMNIST & CIFAR-10 & Shakespeare \\\hline
model $1$ & $74.80\pm2.75$ & $89.25\pm0.74$ & $45.31\pm3.20$ \\
model $2$ & $78.06\pm3.00$ & $90.96\pm0.84$ & $45.83\pm2.48$ \\
model $3$ & $77.85\pm2.90$ & $90.67\pm0.47$ & $46.01\pm2.72$ \\
model $4$ & $78.54\pm2.92$ & $90.45\pm0.54$ & $42.55\pm5.12$ \\\hdashline
auto-tuning & $78.52\pm2.64$ & $89.76\pm0.90$ & $44.71\pm1.12$\\\Hline
\end{tabular}
\end{table}

\begin{table}[!t]
\centering
\caption{Comparison of test accuracy when removing each optimization technique of FedMe. MT, MC, and DML indicate model tuning, model clustering, and deep mutual learning, respectively. %$\cmark$ indicates that FedMe uses the optimization techniques. The number in MC columns indicates the fixed number of clusters.
}
\vspace{-3mm}
\label{tab:result_component}
{\small
\begin{tabular}{ccc|ccc}\Hline
        MT & DML & MC & FEMNIST & CIFAR-10 & Shakespeare \\ \hline
         &  &  & 75.85$\pm$3.27 & 88.19$\pm$0.53 & 37.27$\pm$3.42 \\ 
        \cmark &  &  & 76.46$\pm$3.28 & 89.76$\pm$1.21 & 45.59$\pm$4.04 \\ 
         & \cmark &  & 76.29$\pm$4.22 & 86.92$\pm$2.63 & 42.95$\pm$5.05 \\ 
         %& \cmark & 2 & 74.71$\pm$4.22 & 87.47$\pm$2.12 & 39.63$\pm$1.92 \\ 
         &  & \cmark & 75.87$\pm$3.60 & 88.14$\pm$0.93 & 36.86$\pm$2.54 \\ 
         & \cmark & \cmark & 76.13$\pm$3.71 & 86.73$\pm$2.54 & 43.30$\pm$1.24 \\ 
        \cmark & \cmark &  & 78.76$\pm$2.26 & 89.67$\pm$0.87 & 45.22$\pm$3.70 \\
        %\cmark & \cmark & 2 & 77.94$\pm$3.40 & 89.86$\pm$0.94 & 44.99$\pm$1.27 \\ 
        \cmark &  & \cmark & 77.64$\pm$3.52 & 89.79$\pm$1.23 & 46.32$\pm$3.33 \\ \hdashline
        \cmark & \cmark & \cmark & 78.52$\pm$2.64 & 89.76$\pm$0.90 & 44.71$\pm$1.12 \\ \Hline
\end{tabular}
}
\end{table}

\begin{table*}[!t]
\caption{Impact of data heterogeneity and fine-tuning. }
\vspace{-3mm}
\label{tab:data_hetero}
\begin{tabular}{lcccc}\Hline
 & IID & $\alpha_{label}=5$ & $\alpha_{label}=0.5$ & $\alpha_{label}=0.1$ \\\hline
Centralized w/o fine-tuning& $86.50\pm0.59$ & $85.67\pm0.67$ & $85.96\pm0.54$ & $86.20\pm0.95$ \\
Centralized w/ fine-tuning & $86.45\pm0.34$ & $87.19\pm0.56$ & $90.80\pm0.92$ & $95.36\pm1.16$ \\\hdashline
FedAvg w/o fine-tuning & $87.00\pm0.30$ & $86.64\pm0.34$ & $86.05\pm0.48$ & $80.86\pm1.95$ \\
FedAvg w/ fine-tuning & $85.91\pm0.67$ & $86.40\pm0.84$ & $89.59\pm0.94$ & $\textbf{94.60}\pm1.12$ \\
HypCluster w/o fine-tuning & $\textbf{87.10}\pm0.30$ & $85.21\pm0.51$ & $85.21\pm1.22$ & $82.43\pm1.23$ \\
HypCluster w/ fine-tuning & $86.03\pm0.60$ & $84.45\pm0.48$ & $88.54\pm1.42$ & $94.03\pm1.42$ \\
pFedMe & $33.14\pm31.71$ & $70.13\pm0.87$ & $79.46\pm2.08$ & $86.09\pm3.38$ \\
FedMe w/o fine-tuning & $85.78\pm1.17$ & $86.13\pm1.07$ & $88.15\pm0.52$ & $85.22\pm2.92$ \\
FedMe w/ fine-tuning & $86.04\pm0.99$ & $\textbf{86.94}\pm1.15$ & $\textbf{90.96}\pm0.84$ & $94.18\pm1.24$\\\Hline
\end{tabular}
\end{table*}

\subsubsection{Q3. What techniques of FedMe impact to the inference accuracy?}
We investigate the impact of optimization techniques of FedMe to the accuracy.
FedMe uses the three optimization techniques; model tuning (MT), deep mutual learning (DML), and model clustering (MC).
%We remove model replacement, deep mutual learning, and model clustering, respectively, and conduct experiments.
Table~\ref{tab:result_component} shows the results that FedMe either partially or fully uses optimization techniques.
%Recall that FedMe gradually increases the number $K$ of clusters from 1 to 4.
%Thus, in the model clustering, we check two patterns; $1$ indicates that FedMe does not use model clustering, and $2$ indicates that FedMe fixes the number of clusters as two. 
%We here note that when we set $K$ as larger than 2, the accuracy is less than that of $K=2$.
% , where curriculum means an increase in the number of clusters step by step, and w/o curriculum ($K=1$) and w/o curriculum ($K=2$) means the case without model clustering and with $2$ clusters constantly, respectively.

From Table~\ref{tab:result_component}, we can see that the accuracy of FedMe with all optimization techniques is higher than that without all techniques. 
We first see how much the accuracy improves when FedMe uses a single optimization technique.
The model tuning has the most impact among the optimization techniques in all setting.
%Clients automatically tune their model architectures to the different ones. 
Since they use accurate models more than initial models, the accuracy improves.
Deep mutual learning also improves the accuracy except for CIFAR-10.
The result indicates that deep mutual learning is effective in mutually learning personalized and exchanged models for leveraging predictions of models. 
Different from model tuning and deep mutual learning, model clustering does not improve the accuracy.
We design the model clustering to combine deep mutual learning, so the model clustering itself is not effective.

Next, we investigate the combinations of optimization techniques.
The accuracy of FedMe with two optimization techniques is generally higher than that of FedMe with a single optimization technique.
The result indicates that each optimization technique has effective interaction to improve the accuracy.
For example, in Shakespeare, FedMe with deep mutual learning and model clustering achieves higher accuracy than that with deep mutual learning though model clustering itself is not effective.
While, some combinations decrease the accuracy, for example, DML+MC in FEMNIST and MT and DML in CIFAR-10.
This deterioration is caused by ineffectiveness of model tuning and model clustering methods. 
We have research opportunities to improve the accuracy more, so we remain these tasks as our future work.

\subsubsection{Q4.How fast is the learning process of FedMe compared with the state-of-the-art methods?} 
\begin{figure}[!t]
 \centering
    \begin{minipage}[t]{1.0\linewidth}
        \centering
        \includegraphics[width=0.80\linewidth]{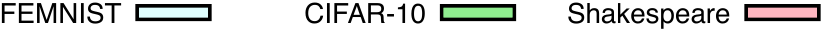}
    \end{minipage}
    \\
    \begin{minipage}[t]{0.49\linewidth}
        \centering
        \includegraphics[width=1.0\linewidth]{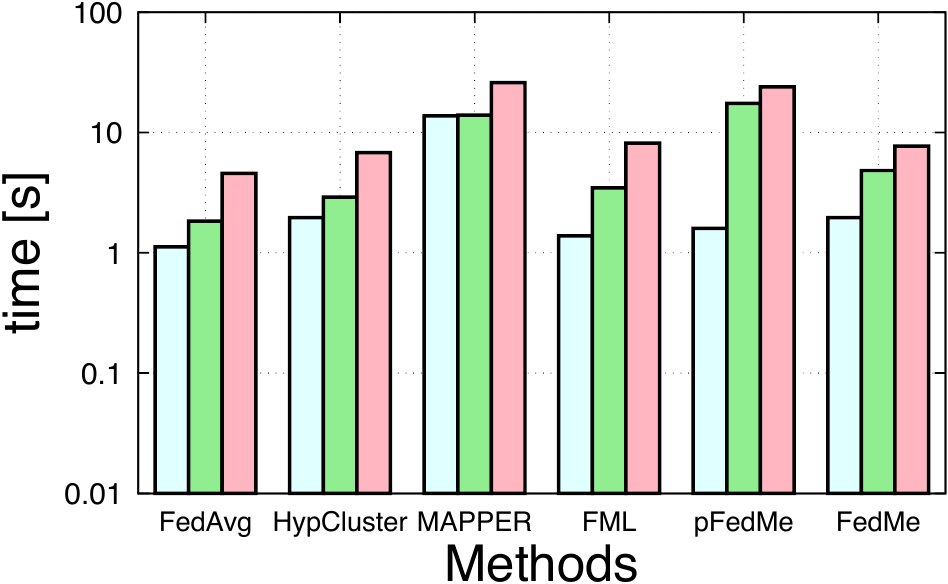}
        \subcaption{Client}
        \label{fig:server_time}
    \end{minipage}
    \begin{minipage}[t]{0.49\linewidth}
        \centering
        \includegraphics[width=1.0\linewidth]{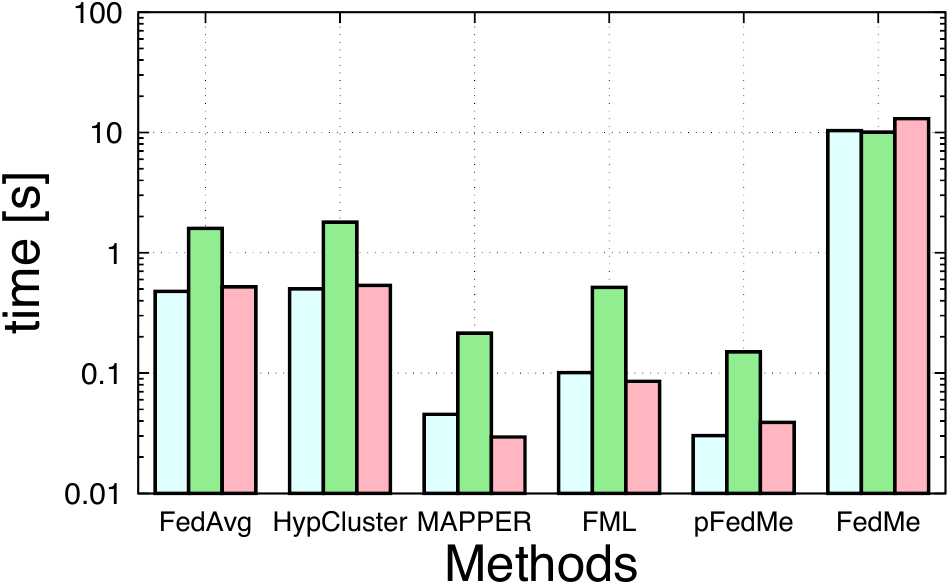}
        \subcaption{Server}
        \label{fig:client_time}
    \end{minipage}
\caption{Run time per global communication round.}
\label{fig:result_runtime}
\end{figure}

We evaluate run time on training phase in each method. Figure~\ref{fig:result_runtime} shows the 
average run time of client and server process on global communication rounds. 
From this result, we can see that FedMe's running time on clients is competitive with other methods, 
though FedMe trains two models. On the other hand, FedMe's running time on server is larger than other methods because the server on FedMe uses model clustering after obtaining the outputs of all the personalized models. This is a time-consuming task compared with other methods. Note that we use the same hardware for the server and the clients in our experiments, while the server generally has more powerful computing resources in real-world scenarios. This means the computation cost on the server tends to be smaller in real-world applications. In addition, we can also control the computation cost of the server by changing the size of unlabeled data. Thus, seeing the accuracy gain of FedMe, we believe that the computation cost on the server is affordable. 
%We can confirm that FedMe efficiently and accurately builds the personalized models.

%  \begin{figure}[!t]
%  \centering
%     \begin{minipage}[t]{1.0\linewidth}
%         \centering
%         \includegraphics[width=0.80\linewidth]{images/legend_data_time.pdf}
%     \end{minipage}
%     \\
%     \begin{minipage}[t]{0.49\linewidth}
%         \centering
%         \includegraphics[width=1.0\linewidth]{images/client_time.pdf}
%         \subcaption{Client}
%         \label{fig:server_time}
%     \end{minipage}
%     \begin{minipage}[t]{0.49\linewidth}
%         \centering
%         \includegraphics[width=1.0\linewidth]{images/server_time.pdf}
%         \subcaption{Server}
%         \label{fig:client_time}
%     \end{minipage}
% \caption{Run time per global communication round}
% \label{fig:result_runtime}
% \end{figure}

\subsubsection{Q5. What is the impact of data heterogeneity and fine-tuning on FedMe?}
\label{sec:experiment_Q4}
We finally investigate the impact of data heterogeneity. 
We conduct experiments using the CIFAR-10 dataset varying the degree of data heterogeneity controlled by $\alpha_{label}$.
The smaller $\alpha_{label}$ indicates greater data heterogeneity, and IID indicates that data distribution and the size of local data on clients are the same. 
% in four environments: $\alpha_{label}$, which represents the degree of data heterogeneity, is varied to $5, 0.5, and 0.1$, and IID, in which the data distribution and size were both uniforms. Here, the smaller $\alpha_{label}$ is, the more substantial the heterogeneity.
We compare FedMe with Centralized and  the top three most accurate existing methods in Table~\ref{tab:result_baseline}, FedAvg, HypCluster, and pFedMe.
Table~\ref{tab:data_hetero} shows the accuracy of each method with and without fine-tuning.
First, FedAvg and HypCluster are equally accurate and have the highest accuracy in IID.
The result indicates that when data is distributed in IID, it is enough to average model parameters of models on clients.
We can also see that fine-tuning is not effective in IID.
As $\alpha_{label}$ decreases (i.e., data heterogeneity becomes greater), the accuracy of methods without fine-tuning decreases, but that of methods with fine-tuning increases.
This is because labels of local data have skews, training and test datasets have similar labels.
The accuracy of pFedMe, which is the personalized federated learning method, also increases as $\alpha_{label}$ decreases. 
However, pFedMe always reports the worst performance (compared to other fine-tuning methods) even though it was designed as a solution to high degree of data heterogeneity and includes techniques similar to fine-tuning. 
%This result indicates that data heterogeneity can be solved by fine-tuning. 
This result was not observed in previous studies. 
For FedMe, when equipped with fine-tuning, it is the best (or the runner-up) method when local data is not IID. 
FedMe without fine-tuning is also the best (or the runner-up) method among the methods without fine-tuning. 
These results show that FedMe works well for high degree of heterogeneity and demonstrate the robustness of FedMe when fine-tuning is absent.

\section{Conclusion and Future Work}
\label{sec:6_conclusion}
In this paper, we presented FedMe, a novel federated learning method that builds personalized models with automatic model architecture tuning.
In FedMe, clients exchange their models to tune and train their personalized models.
FedMe can train models with different architectures by exchanging models and deep mutual learning.
Our experiments showed that FedMe is more accurate than the state-of-the-art methods and can automatically tune the model architecture.
%In addition, we show that each component of FedMe works effectively.
%We finally explore how the heterogeneity of the data affects the accuracy of each method and show that FedMe is more robust to the degree of heterogeneity than the other methods and that fine-tune can sufficiently create personalized models when the heterogeneity level is high.

As our future work, we plan to extend model tuning and model clustering methods for tuning model architecture more flexibly.
Although FedMe automatically tunes the model architecture, the candidates are only model architectures that clients design in advance. Thus, FedMe may not work well if optimal model architectures are not designed.
We can improve model tuning methods for tuning models more flexibly, such as network architecture search.
Additionally, model clustering is not effective much, so we can extend it to improve the accuracy.

%We plan to search for more flexible model architecture, such as the general network architecture search (NAS), while creating personalized models.

%\input{7_acknowledgement.tex}

% \begin{acks}
% \input{7_acknowledgement}
% \end{acks}

%\clearpage

\bibliographystyle{abbrv}
\bibliography{bibliography}

\begin{comment}

\appendix

\section{Research Methods}

\subsection{Part One}

Lorem ipsum dolor sit amet, consectetur adipiscing elit. Morbi
malesuada, quam in pulvinar varius, metus nunc fermentum urna, id
sollicitudin purus odio sit amet enim. Aliquam ullamcorper eu ipsum
vel mollis. Curabitur quis dictum nisl. Phasellus vel semper risus, et
lacinia dolor. Integer ultricies commodo sem nec semper.

\subsection{Part Two}

Etiam commodo feugiat nisl pulvinar pellentesque. Etiam auctor sodales
ligula, non varius nibh pulvinar semper. Suspendisse nec lectus non
ipsum convallis congue hendrerit vitae sapien. Donec at laoreet
eros. Vivamus non purus placerat, scelerisque diam eu, cursus
ante. Etiam aliquam tortor auctor efficitur mattis.

\section{Online Resources}

Nam id fermentum dui. Suspendisse sagittis tortor a nulla mollis, in
pulvinar ex pretium. Sed interdum orci quis metus euismod, et sagittis
enim maximus. Vestibulum gravida massa ut felis suscipit
congue. Quisque mattis elit a risus ultrices commodo venenatis eget
dui. Etiam sagittis eleifend elementum.

Nam interdum magna at lectus dignissim, ac dignissim lorem
rhoncus. Maecenas eu arcu ac neque placerat aliquam. Nunc pulvinar
massa et mattis lacinia.

\end{comment}

\end{document}